\documentclass[letterpaper]{article} 
\usepackage{aaai2026}  
\usepackage{times}  
\usepackage{helvet}  
\usepackage{courier}  
\usepackage[hyphens]{url}  
\usepackage{graphicx} 
\usepackage{natbib}  
\usepackage{caption} 
\usepackage{algorithm}
\usepackage{algorithmic}

\usepackage{newfloat}
\usepackage{listings}
\DeclareCaptionStyle{ruled}{labelfont=normalfont,labelsep=colon,strut=off} 
\floatstyle{ruled}
\newfloat{listing}{tb}{lst}{}
\floatname{listing}{Listing}
\author{
    Ting-Hao `Kenneth' Huang\textsuperscript{\rm 1},~
    Ryan A. Rossi\textsuperscript{\rm 2},~
    Sungchul Kim\textsuperscript{\rm 2},~
    Tong Yu\textsuperscript{\rm 2},\\
    Ting-Yao `Edward' Hsu\textsuperscript{\rm 1},~
    Ho Yin Ng (Sam)\textsuperscript{\rm 1},~
    Clyde Lee Giles\textsuperscript{\rm 1}
}
\affiliations{
    \textsuperscript{\rm 1}The Pennsylvania State University, 201 Old Main, University Park, PA, USA\\
    \textsuperscript{\rm 2}Adobe Research, San Jose, CA, USA\\
    \{txh710,~txh357,~sam.ng,~clg20\}@psu.edu\\
    \{ryrossi,~sukim,~tyu\}@adobe.com

}

\usepackage{bibentry}

\usepackage{xspace}
\usepackage{booktabs}
\usepackage{dirtytalk}
\usepackage{cleveref}
\usepackage{multirow}
\usepackage[inline]{enumitem}
\usepackage{ifthen}
\usepackage{xcolor}
\usepackage{subcaption}
\usepackage{comment}
\usepackage{booktabs}

\title{Five Years of \dataset:\\What We Learned and Future Directions for Scientific Figure Captioning}

\newcommand{\dataset}{\textsc{SciCap}\xspace}

\definecolor{ao(english)}{rgb}{0.0, 0.5, 0.0}

\newcommand{\eg}{{\it e.g.}}
\newcommand{\ie}{{\it i.e.}}

\newenvironment{myquotewithindent}%
  {\list{}{\leftmargin=0.1in\rightmargin=0.1in}\item[]}%
  {\endlist}

\newcommand{\participant}[1]{%
    \ifnum#1=1 P001\xspace \else
    \ifnum#1=2 P002\xspace \else
    \ifnum#1=4 P003\xspace \else
    \ifnum#1=5 P004\xspace \else
    \ifnum#1=10 P005\xspace \else
    \ifnum#1=11 P006\xspace \else
    \ifnum#1=14 P007\xspace \else
    \ifnum#1=15 P008\xspace \else
    Unknown\xspace
    \fi\fi\fi\fi\fi\fi\fi\fi
}

\begin{document}

\maketitle


\begin{abstract}
Between 2021 and 2025, the \dataset project grew from a small seed-funded idea at The Pennsylvania State University (Penn State) into one of the central efforts shaping the scientific figure-captioning landscape. 
Supported by a Penn State seed grant, Adobe, and the Alfred P. Sloan Foundation, what began as our attempt to test whether domain-specific training, which was successful in text models like SciBERT, could also work for figure captions expanded into a multi-institution collaboration.
Over these five years, we curated, released, and continually updated a large collection of figure–caption pairs from arXiv papers, conducted extensive automatic and human evaluations on both generated and author-written captions, navigated the rapid rise of large language models (LLMs), launched annual challenges, and built interactive systems that help scientists write better captions.
In this piece, we look back at the first five years of \dataset and summarize the key technical and methodological lessons we learned. 
We then outline five major unsolved challenges and propose directions for the next phase of research in scientific figure captioning.

\end{abstract}

\section{2021-2025: The First Five Years of \dataset}
A scholar once shared a piece of wisdom: a good researcher either ``makes something big out of something small'' or ``makes something small out of something big.''
In other words, one can choose a narrow, well-scoped question and explore it deeply, or take on an ambitious, moonshot idea by starting from its simplest form. 
The first five years of \dataset belong to the former path. 
We chose a laser-focused, narrowly-scoped problem---\textbf{how to generate and evaluate scientific figure captions}---and pursued it with persistence and depth. 
What followed was a journey that began with a small idea and grew into a sustained, multi-institution collaboration that continues today.


This section reviews this journey from 2021 to 2025.


\subsection{Origin Story of \dataset: 133k+ Figure-Caption Pairs from arXiv Papers}
The story of \dataset began in late 2019, when we at Penn State started discussing whether we could build a large-scale dataset of real-world scientific figures and captions to support multimodal modeling for scientific visuals. 
The spark came when Lee walked into Kenneth's office and asked, \textit{``Can we build a CaptionBERT?''}
Lee knew Kenneth had worked on vision-and-language datasets and models before, including the Visual Storytelling (VIST) dataset, and around that time domain-specific BERT variants, such as SciBERT~\cite{beltagy-etal-2019-scibert} and BioBERT~\cite{10.1093/bioinformatics/btz682}, were showing clear advantages.
Their success made us wonder whether scientific figure captions might also benefit from domain-specific modeling, if only we could gather enough real data.

We soon drafted a seed-grant proposal to Penn State outlining two motivations:
First, figure captions in scientific papers were often poorly written, frequently lacking the detail or clarity needed for readers to understand a figure without returning to the full text. 
This seemed like an ideal place for automated assistance. 
Second, if we wanted to build technology to improve caption quality, there was a fundamental data gap: no large-scale, real-world dataset of scientific figures paired with their captions existed. 
Most prior datasets were either synthetic or too small to support reliable captioning models~\cite{Kafle_2018_CVPR,kahou2017figureqa,10.1145/3341162.3345601,Chen_2020_WACV}.
We aimed to fix both problems at once.

The proposal was funded. 
By 2021, we had constructed and released the first version of the \dataset dataset~\cite{hsu-etal-2021-scicap-generating}, focusing on \textbf{graph plots from computer-science and machine-learning arXiv papers.} 
We gathered 295,028 papers published between 2010 and 2020, extracted over 2.17 million figures, identified 416,804 graph plots (19.2\%), and after filtering out compound figures, produced a clean set of \textbf{133,543 single-panel figures}.

In the 2021 \dataset release paper~\cite{hsu-etal-2021-scicap-generating}, we also evaluated several baseline vision-to-language models. 
Their performance was strikingly poor: captions were often low-quality, inaccurate, or even outright gibberish.
The results confirmed our intuition: scientific figure captioning was far from a solved problem, and real progress would require both better data and better models.

\subsection{A Highly Contextual Task: Captioning as Contextual Summary, Not Just Visual Description}
Soon after we released the first version of \dataset, we learned that Adobe Research had also been exploring scientific figure captioning~\cite{10.1145/3442381.3449923}. 
We reached out, and after a few very encouraging conversations, our two groups decided to join forces in Fall 2021. 
That collaboration ended up shaping the direction of \dataset for years.

As we examined the low-quality outputs from our initial models, it became clear that scientific figure captioning is fundamentally a context-rich problem. 
In one of our early meetings, an Adobe collaborator asked a question that stuck with us: \textit{``Where does the information in a caption actually come from?''}
No one, not even a human expert, can write a good caption from the image alone.

From 2021 to early 2022---before ChatGPT or multimodal LLMs were publicly available---we worked together to figure out how to incorporate document context, especially the paragraphs that mention each figure (\eg, ``Figure~3 shows...''), into the captioning process.
However, building effective multimodal models proved difficult, and early attempts to combine images with text yielded only modest improvements.

Gradually, we noticed a pattern: many scientific figure captions were essentially paraphrases or compressed summaries of the paper's own descriptions. 
That observation led to a simple but surprising experiment. 
We treated figure captioning as a text-summarization problem and fine-tuned a summarization model, \textsc{Pegasus}~\cite{zhang2020pegasus}, using only the figure-mentioning paragraphs---no images at all. 
The text-only approach worked remarkably well. 
We also found that roughly 75\% of caption words could be traced directly to those paragraphs or the figure's OCR.

This work resulted in our INLG 2023 paper, which received the Best Long Paper award and was selected as one of the three finalists for the Best Evaluation Paper award~\cite{huang-etal-2023-summaries}.
It demonstrated that scientific figure captioning behaves much more like contextual summarization than visual description. 
Furthermore, a small human evaluation in the paper confirmed our motivating observation:
many author-written captions were mediocre at best.
PhD students rated over 50\% of author-written captions as \textit{unhelpful}.
This result highlights the limitations of caption authoring practices and the potential for LLM assistance to improve caption utility for readers.

\subsection{LLMs Judge Caption Quality Like PhD Students, Not Undergraduates}
Then came ChatGPT in late 2022, a history-defining moment. 
For the first time, fluent, high-quality text generation became accessible to millions. 
After some quick empirical testing, we made a strategic decision: instead of building new architectures specifically for scientific figure captioning, we would focus on understanding and leveraging LLMs' ability to generate captions for scientific figures.
Our goal is to help authors and readers work with clearer, more useful captions, and it became clear that building on LLMs would be a more effective and generalizable path than creating application-specific models.


Once we reached some major milestones in caption generation, the next obvious challenge was evaluation. 
Our INLG paper had already reminded us of a familiar truth echoed across many prior text-generation projects: automatic evaluation is not enough. 
Traditional reference-based metrics like BLEU and ROUGE did not correlate reliably with human judgments. 
And human evaluation (at least for scientific figure captions) was expensive and logistically difficult. 
Scientific captions require expert background knowledge; generic online crowd workers cannot evaluate captions for physics or electrical-engineering figures, and even NLP researchers are often unqualified outside their own subfields.

Against this backdrop, as well as with ChatGPT reshaping the field, we began experimenting with LLMs as automatic judges in early 2023. 
PhD students and trained undergraduates rated captions generated by different approaches, and we compared their ratings to those produced by LLMs. 
The results, published at EMNLP 2023~\cite{hsu-etal-2023-gpt}, were interesting: 
LLM judgments aligned well with PhD students' ratings. 
Undergraduate ratings, however, diverged. 
Less-experienced readers cared more about clear take-away messages, while experts prioritized numerical precision and technical detail. 

Our study landed at the early wave of the ``LLM-as-judge'' trend and drew more attention than we expected. 
Interestingly, a later 2024 study would report that LLM judges sometimes favor their own generations~\cite{10.5555/3737916.3740113}.
Our work avoided that bias by deliberately excluding LLM-generated captions from the evaluation pool.

\subsection{\dataset Challenge: Building a Community and Expanding to 476k+ Figure-Caption Pairs}
\begin{figure}[t]
    \centering
    \includegraphics[width=\linewidth]{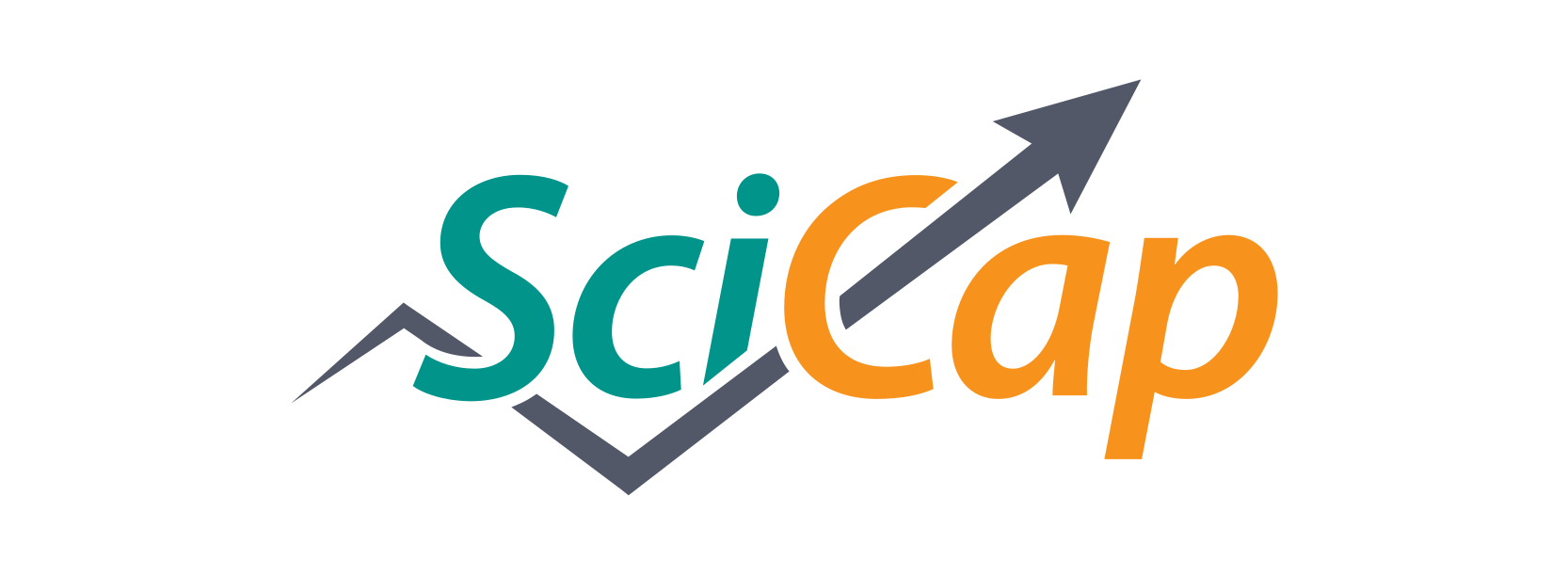}
	  \caption{The \dataset logo, designed in 2023 for the first \dataset Challenge and used across all related projects since.}
\label{fig:logo}
\end{figure}

We launched the first \dataset Challenge in 2023 (Figure~\ref{fig:logo}) at the ICCV workshop Closing the Loop Between Vision and Language. 
We invited teams around the world to generate captions for an expanded version of our dataset, \ie, the \dataset Challenge dataset, which contained \textbf{476,389 single-panel scientific figures across 8 scholarly domains and 5 figure types}.\footnote{The \dataset Challenge dataset~\cite{hsu2025large} included figures from arXiv papers across computer science, physics, math, economics, statistics, quantitative biology, quantitative finance, and electrical engineering. Figures were categorized into node diagrams, equations, graph plots, scatterplots, and bar charts. We limited the per-paper sampling to encourage participation from teams with modest computational budgets. This dataset was more than three times the size of our original 2021 \dataset release.}
Each figure came with its image, OCR text, figure type, and the paragraphs in the arXiv that explicitly mentioned it. 
When we designed the specification of the \dataset Challenge, we intentionally made the figure-mentioning paragraphs part of the standard input that every team received. 
We wanted this design choice to send a clear signal: scientific figure captioning is a highly context-dependent generation task; a good caption cannot be written from an image alone. 

Built on the outputs submitted to the 2023 Challenge, we conducted a large-scale human and automatic evaluation study, later published at TACL~\cite{hsu2025large} and presented at NAACL 2025. 
In this paper, we found that multimodal LLMs (MLLMs), particularly GPT-4V, consistently produced captions preferred by professional editors over all other approaches, including human-written captions. 
This was true across domains, across figure types, and even for papers published after the models' claimed training cut-off date.
In cases where authors wrote captions carelessly, MLLM-generated captions were often clearly better---a humbling result, but also an exciting one for the potential of assistive tools.

Since 2023, we have continued to run the \dataset Challenge annually---at IJCAI 2024 and at the LM4Sci Workshop at COLM 2025. 
Over time, this effort has helped form a small yet active community working on scientific figure captioning. 
Some teams have even returned to participate again after their first participation; we even collaborated with a team to publish their approach~\cite{10.1007/978-981-96-8912-5_6}.
Our very first \dataset paper~\cite{hsu-etal-2021-scicap-generating}, which originally was a Findings paper without fancy modeling, has been cited over 130 times. 

\subsubsection{Constant Trade-offs Between Automatic and Human Evaluation.}
Each year we ran the \dataset Challenge, we faced the same dilemma: automatic evaluation is cheap and fast but unreliable, while human evaluation is reliable but extremely slow and expensive. 
Across the 2023-2025 challenges, we experimented with different ways to balance this trade-off. 
In 2023, we gave two awards: one based on \texttt{ROUGE-2-Normalized} scores on the full hidden test set, and another on a curated subset of higher-quality captions, motivated by our finding that models performing well overall could still fail on well-written captions~\cite{huang-etal-2023-summaries}.
However, rankings barely changed across the two sets.
In 2024, we switched to human evaluation on a small sample of 200 figures and split the challenge into two tracks: a short-caption track, where systems had to produce captions no longer than the originals for at least 30\% of samples, and a long-caption track with the opposite constraint. 
This time, the results diverged: teams outperformed the original authors of captions in the long-caption track, but authors still performed better in the short-caption track. 
This could be due to humans' known preference for longer captions.
In 2025, to mitigate this bias, we went with two distinct tracks, one judged by humans and one judged automatically, using the new \textsc{LaMP-Cap} dataset~\cite{ng2025lamp} focused on personalized captioning.
This time, the results diverged again across two sets.
Teams were able to mimic the author's style per automatic evaluations, but this often hurt caption quality perceived by humans.
In fact, human evaluators consistently preferred GPT-4o's vanilla captions over personalized ones. 


If these three years taught us anything, it is that we still understand very little about how to reliably measure the quality of scientific figure captions \textit{at scale}.
This particular type of captioning is highly contextual, often requires specialized knowledge to assess, and even author-written captions are not always reliable.

\subsection{Supporting Caption Writers: Promising, but Highly Nuanced}
As our data, generation methods, and evaluation practices began to mature, we turned to the part of the loop that matters most for real impact: helping writers. 
We focused on the writer side because readers have little control over caption quality---most readers cannot revise captions or run AI models---but improving the writing process can immediately benefit thousands of readers at once. 

Our first step was building \textsc{SciCapenter}, an interactive caption-writing assistant that displays the figure, its figure-mentioning paragraphs, and several AI-generated draft captions, and allows writers to iteratively revise them with quality ratings and contextual guidance~\cite{10.1145/3613905.3650738}.
A pilot study suggested that these suggestions reduced cognitive load and helped writers produce higher-quality text. 
We then conducted a separate, deeper study in which 18 researchers rewrote captions from their own recently published papers while using multiple LLM-generated captions as starting points~\cite{ngunderstanding}.
Through video-based interaction analysis, we found that writers typically began by copying or adapting AI-generated captions, preferred longer and detail-rich drafts, and relied heavily on the textual–visual alignment of the suggestions. 
However, they struggled when figures were conceptually complex or when AI suggestions conflicted with disciplinary norms.
We also saw broader themes~\cite{ng-etal-2025-understanding}: caption writing is deeply multimodal, writers often lack linguistic confidence but trust their domain expertise, they benefit from multiple diverse AI suggestions, and caption norms vary dramatically across fields.



\subsection{Personalized Captioning Through Multimodal Profiles}
In our caption-writing studies~\cite{ngunderstanding,ng-etal-2025-understanding}, we noticed a recurring gap: even when AI-generated captions were of fine quality, authors often felt they were too generic and did not match their writing style or the stylistic conventions of their target venues. 
This pushed us to think seriously about personalization. 
Although personalized text generation is now common with LLMs~\cite{zhang2024personalization}, the key practical question for our task was: \textit{what information can we use to characterize the style of a group of co-authors?} 
Using all papers written by each co-author was an obvious answer, but it was expensive and conceptually messy, since scholarly writing is collaborative and we rarely know which author actually wrote which caption. 
So we turned to something more grounded and reliable: information within the same paper. 

This led us to create \textsc{LaMP-Cap}~\cite{ng2025lamp}, a dataset derived from the \dataset Challenge dataset and designed specifically for personalized scientific figure captioning using multimodal profiles.
For each target figure, \textsc{LAMP-CAP} provides the figure image and its figure-mentioning paragraphs as model input, along with up to three additional figures from the same paper (each with its own image, caption, and paragraphs) as a style profile. 
Across 110,828 target figures, we tested four LLMs and consistently found that adding these multimodal profiles helped models generate captions closer to the original author-written ones.
Human evaluation further confirmed the usefulness of personalization.
Interestingly, human judges preferred captions generated with one profile figure over both no-profile and all-profile conditions. 
This echoes a common dilemma in personalization~\cite{zhang2024personalization}: a little tailoring improves quality and feels helpful to users, but pushing too hard to mimic the user's style can hurt the generation quality.

\section{Challenges and Future Directions}
In this position paper, we argue that scientific figure captioning is a clear example of how a technology that appears powerful---LLM-generated captions for scientific figures---still faces many obstacles before it can be truly useful in real-world settings. 
Working on this problem has shown us how much depth and complexity can surface when we study real tasks that look small and well defined from the outside. 
It also shows a major opportunity for the technical HCI community, because many of the hardest remaining challenges in scientific figure captioning are fundamentally \textit{human-centered}.

In the following, we describe the challenges and potential future directions that emerged from our five years of experience working on the \dataset project and related efforts.
These ideas grew out of ongoing discussions and continuous reflection within our team, and we believe sharing them can benefit the broader research community.



\subsection{Challenge \#1: Captioning with Incomplete or Missing Context}

Real-world writing rarely unfolds in neat, fully drafted documents.
To help authors, captioning systems must handle partial or missing context for figures in unfinished papers or for standalone visuals circulating on social media, in news, or in advertisements. 
\textbf{18.81\% of figures in the original \dataset dataset lacked any identified mentions}~\cite{huang-etal-2023-summaries}.
Generating meaningful captions when only the image is available remains one of the field's most practical yet unsolved problems.

\subsection{Challenge \#2: Captions for a Variety of Readers}
\dataset's evaluations showed that reader expertise shapes caption preference~\cite{hsu-etal-2023-gpt}.
Undergraduates valued clear takeaways; PhD students cared about methodological precision and numbers. 
Future models might adapt captions for different audiences, such as experts, students, and practitioners, which will require advances in audience modeling, controllable generation, and implicit preference inference. 
Building audience-aware captioning systems without relying on extensive user profiling remains an open challenge in personalization.


\subsection{Challenge \#3: Personalizing to the Writer's Style}
The experimental results from both the \textsc{LaMP-CAP} dataset~\cite{ng2025lamp} and the \dataset Challenge 2025 showed a clear trade-off: 
strong stylistic imitation can reduce factual accuracy or clarity. 
This tension between stylistic faithfulness and communicative quality raises important questions about authorship, identity, and the design of human-AI co-writing systems. 
It also brings up a broader question that some NLP researchers have asked: to support writing, do we actually need personalized text generation, or is generic generation sufficient for most authors to build on?

\subsection{Challenge \#4: Studying and Supporting the Writing Process}
Most caption research evaluates outputs from the reader's perspective, not the writer's.
Yet to understand how AI truly helps scientists, we must study caption writing in realistic conditions. 
Recruiting authors actively drafting their own papers is difficult; proxy studies, \eg, rewriting old captions or captioning others' figures~\cite{ngunderstanding,ng-etal-2025-understanding}, only approximate the real task. 
The Sloan-funded deployment of \textsc{SciCapenter} opens a rare opportunity to observe genuine writing-in-progress~\cite{Huang2024_ScientificCaptions}, but conducting such ecologically valid studies remains expensive and methodologically hard.

\subsection{Challenge \#5: Bridging Captioning and Figure Understanding}
Despite shared goals, figure captioning and figure understanding research have mostly evolved in isolation. 
Work on figure parsing, chart reasoning, and figure QA rarely intersects with user-facing captioning tasks that need evaluation beyond parsing accuracy. 
Conversely, many captioning projects, including our own, did not fully evaluate faithfulness or correctness with respect to the visual content.
Captioning focuses on communication for human readers, while figure understanding is often developed as a reasoning benchmark.
From a captioning perspective, our approach to the task treated it as writing support, which allowed us to worry less about strict grounding because authors remained in the loop. 
Still, there are clear scenarios where visual faithfulness matters.
Bringing these threads together could yield more holistic systems that both understand and explain scientific visuals, reduce duplication of effort, and push the field toward genuine multimodal comprehension.


\section{Conclusion}
Scientific figure captioning is a small, well-scoped task that turned out to be far more complex than we expected. 
Working on it taught us practical lessons about text generation, evaluation, and how AI-generated text can fit into real writing workflows. 
There is still much to explore, and we see this work as the starting point for the next phase of research on captioning and multimodal language technologies.


\section{Acknowledgments}
We thank all co-authors and collaborators on \dataset-related projects, as well as every team that participated in the \dataset Challenges. 
This work has been supported by a Penn State IST Seed Grant (2020), Adobe Research's gift (2022), and the Alfred P. Sloan Foundation grant (2024-2026, \#G-2024-22721).


\bibliography{bib/aaai2026}

\end{document}